%% file: main.tex
\documentclass[twoside]{article}
\usepackage{natbib}
\usepackage{graphicx}
\usepackage{bm}
\usepackage{tikz}
\usepackage{amsmath}
\usepackage{amssymb}
\usepackage{multirow}
\usepackage{subcaption}
\usetikzlibrary{calc,fit,positioning}

\usepackage[colorinlistoftodos,prependcaption,textsize=tiny,disable]{todonotes}

\usepackage{tikz}
\usetikzlibrary{shapes,arrows} 
\usetikzlibrary{positioning}
\usetikzlibrary{bayesnet}
\usetikzlibrary{patterns}
\usetikzlibrary{backgrounds}

\usepackage{adjustbox}

\usepackage[accepted]{aistats2020}
\begin{document}

%

%

\twocolumn[

\aistatstitle{Bayesian nonparametric shared multi-sequence time series segmentation}

\aistatsauthor{ Olga Mikheeva \And Ieva Kazlauskaite \And Hedvig Kjellstr\"{o}m \And Carl Henrik Ek}

\aistatsaddress{ KTH Royal Institute \\of Technology \\Stockholm, Sweden\\\texttt{olgamik@kth.se} \And  University of Bath\\Bath, United Kingdom\\\texttt{I.Kazlauskaite@bath.ac.uk} \And KTH Royal Institute \\of Technology \\Stockholm, Sweden\\\texttt{hedvig@kth.se} \And University of Bristol\\Bristol, United Kingdom\\\texttt{carlhenrik.ek@bristol.ac.uk}} ]

\begin{abstract}
    In this paper, we introduce a method for segmenting time series data using tools from Bayesian nonparametrics. We consider the task of temporal segmentation of a set of time series data into representative stationary segments. We use Gaussian process (GP) priors to impose our knowledge about the characteristics of the underlying stationary segments, and use a nonparametric distribution to partition the sequences into such segments, formulated in terms of a prior distribution on segment length. Given the segmentation, the model can be viewed as a variant of a Gaussian mixture model where the mixture components are described using the covariance function of a GP. 
    We demonstrate the effectiveness of our model on synthetic data as well as on real time-series data of heartbeats where the task is to segment the indicative types of beats and to classify the heartbeat recordings into classes that correspond to healthy and abnormal heart sounds.
\end{abstract}

\section{Introduction}

For interpretable analysis of multiple time series of locally stationary data, it can be beneficial to describe time series data as a sequence of representative stationary segments. For example, human daily activity can be described as a sequence of typical activities: walking, walking up/down, sitting, standing, cycling, etc. Different individuals will have different distribution of activities, while motion during the same activity will look similar across individuals. Another application is medical diagnostics from behavioral patterns, for example, diagnostics of depression or cognitive decline from the observed behavior of the patient. In medical applications, interpretability of the underlying representation is crucial. Temporal segmentation into typical functional behaviors can give an interpretable lower dimensional representation of such data, on which later classification models can be trained. Those applications motivate our current work on sequence segmentation.

In this work, we propose a generative probabilistic model for locally stationary time-series data. The model includes explicit prior on length of segments. Gaussian processes (GPs) are used as priors over functions on each segment. Inference in the model is performed via hybrid variational expectation maximization (EM).

The proposed model and the inference scheme allow for unsupervised Bayesian sequence segmentation. Taking the Bayesian approach allows for incorporating prior information about length of the segments and local behaviors. This is particularly important, since unsupervised segmentation is not a well-constrained problem, and there are many valid segmentation options.

\section{Related Work}

Previous work on locally stationary time series is either focused on fitting a function to data for prediction purposes (\cite{rasmussen2002infinite}) or on change point detection. Change point detection refers to a class of methods for detecting points in a sequence where its characteristics change abruptly~\citep{Aminikhanghahi:2017}. The problem is typically formulated using a cost function, which detects shifts in statistical quantities (such as the mean, the scale, the linear relationship between dimensions, etc.) or a change in the distribution (\emph{e.g.} when a kernel method is used to describe the data), and a penalty term (that allows to impose constraints such a fixed number of change points)~\citep{Truong:2018}. Typically, change point detection approaches do not impose any priors on the distribution of the data in each segment of the sequence and they focus on point estimates for the locations of the change points rather than an entire distribution over these locations. Furthermore, these lines of work usually focus on a single sequence. In contrast, we leverage the availability of multiple time series, assuming they share types of locally stationary segments. 

While time series analysis is typically performed in the time domain, some previous work also focused on the representation of the data in the frequency domain. For example, both~\cite{deng2012robust} and~\cite{balili2015classification} use a wavelet decomposition of the signal and analyse the resulting spectrograms to detect the anomalies and to perform segmentation using hand-picked frequency bands. 

There has been work on nonparametric Bayesian methods for segmentation in various domains, including image data~\citep{ghosh2011spatial} and time series, focusing on different aspects of the broader problem of segmentation and clustering. \cite{Dhir:2016} introduced an unsupervised segmentation approach that uses a hierarchical Dirichlet process mixture model for clustering of similar segments, and a Hidden Markov model over an infinite state space to model the observations in each segment. The authors note that it is difficult to specify the domain and the hyper-priors of the model parameters, and propose employing Bayesian optimisation as a black-box optimisation tool that allows to search the space of the parameters given the expensive fitting of the model~\citep{Dhir:2017:2}. Furthermore, the proposed model encourages the creation of redundant states and rapid switching amongst these~\citep{Dhir:2017:2}. Various extensions have been proposed to address this issue~\citep{Fox:2008, Dhir:2017}, however, they introduce additional parameters for the self-transition biases. Another line of work has focused on Bayesian nonparameteric clustering of batch-sequential data without considering segmentation per se. For example,~\cite{Campbell:2013} have utilised dependent Dirichlet process mixture model for the clustering of the segments and a Gaussian process for the modeling of the data within the segments. 

\section{Methodology}
In this paper we focus on the task of segmenting a set of time-series data into a discrete set of sequences. Importantly, we do not know the explicit form of the individual segments nor how they combine to form each time-series. Performing segmentation in such scenario is challenging and can only be done by specifying priors that encodes a preference towards specific structures. In this paper we take a non-parametric approach to both the model of the sub-sequences and the structure of how they combine. In specific, we use Gaussian process priors \citep{Rasmussen:2005} to model each sub-sequence and use a nonparametric prior on partitions, formulated in terms of a distribution on segment length, to describe how they combine. Importantly both these priors allows informative structures to be encoded if known a-priori while at the same time providing support for a large class of solutions if supported by the data.

We will now describe the generative processes specified by our model. First we draw a segmentation of a sequence with specific length. Given a segmentation each sub-sequence is assigned a GP from which a realisation of the sub-sequence can be drawn. Importantly, the model of the sub-sequences are shared across multiple time-series. We will now proceed to write down the model corresponding to the generative proceedure defined above.

\subsection{Model}
Let the data set be $\{\bm{Y^d}, \bm{X^d}\}_{d=1}^D$, where $D$ is the number of sequences, $\bm{Y^d}= \{y^d_n\}_{n=1}^{N_d}$ is the data for sequence $d$, and $\bm{X^d}= \{x^d_n\}_{n=1}^{N_d}$ is the corresponding set of time stamps. 

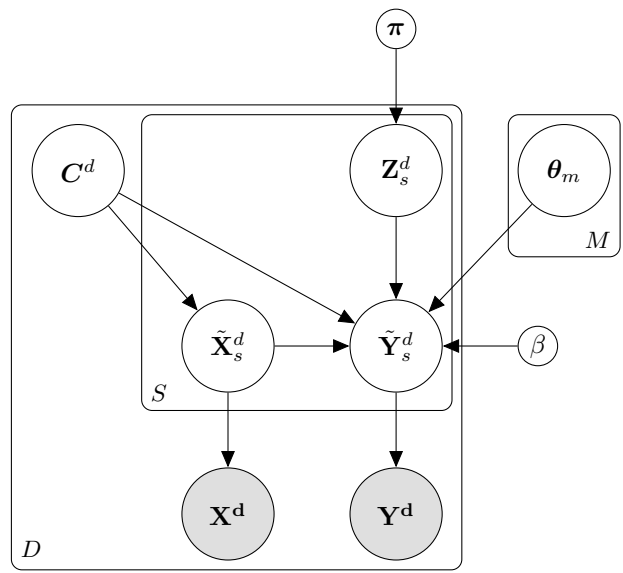
\begin{figure}[ht]
\centering
  \resizebox{0.5\textwidth}{!}{%
    \input{graphical_model.tex}
  }
  \caption{Generative model. Given breaks $C$, segmentation $S(C)$ determines how original sequences are split into shorter segments $\{(Y_s,X_s)\}_{s=1}^{S(C)}$.}
  \label{fig:model_pgm}
\end{figure}

The model is defined as follows:
\begin{equation}
\begin{split}
     p(\bm{Y},&\bm{Z},\bm{C},\bm{\theta},\beta, \bm{\pi}|\bm{X})\\ & =p(\bm{Y}, \bm{C},\bm{Z}|\bm{X}, \bm{\theta},\bm{\pi},\beta) p(\bm{\pi}) p(\bm{\theta})p(\beta), 
\end{split}
\end{equation}
\begin{equation}
\begin{split}
    p(\bm{Y},&\bm{C},\bm{Z}|\bm{X}, \bm{\theta},\bm{\pi},\beta) \\ & =\prod_{d=1}^D p(\bm{Y^d}, \bm{Z^d}| \bm{C^d},\bm{X^d}, \bm{\theta},\bm{\pi}, \beta) p(\bm{C^d}), 
\end{split}
\end{equation}
with $\bm{C}^d =\{c_n^d\}_{n=1}^{N_d}$, where $c_n \in \{0,1\}$ and 1 indicates start of a new segment. The corresponding graphical model is shown in Figure~\ref{fig:model_pgm}.  We denote the number of resulting segmentation by $S(\bm{C^d})$, the number of segments by $S^d$, and the corresponding segment length by $\{l_s^d\}_{s=1}^{S^d}$. Both $S$ and $L$ here are deterministic functions of $C$. The probability of segmentation is then defined as:
\begin{equation}
    p(\bm{C}^d)=\prod_{s=1}^{S^d}p(l_s^d).
\end{equation}

We propose to use $l_s \sim \text{Exp}(\lambda)$, however, other priors may be used within the same framework.

Given the segmentation, the joint probability of $\bm{Y}$ and $\bm{Z}$ can be factorized over the segments:

\begin{equation}
\begin{split}
     p(\bm{Y^d}, & \bm{Z^d}| \bm{C^d},\bm{X^d}, \bm{\theta},\bm{\pi}, \beta) \\
     & =\prod_{s=1}^{S^d}p (\bm{Y^d_s}, \bm{Z^d_s}| \bm{X^d_s}, \bm{\theta},\bm{\pi}, \beta)  \\
     & = \prod_{s=1}^{S^d}p (\bm{Y^d_s}|\bm{Z^d_s},\bm{X^d_s}, \bm{\theta}, \beta)p (\bm{Z^d_s}|\bm{\pi})
\end{split}
\end{equation}
where $\bm{Z^d_s}$ is the latent cluster assignment for the d$^{th}$ sequence and it follows a categorical distribution $\bm{Z^d_s}|\bm{\pi} \sim \text{Cat}(\pi_1,...\pi_M)$, where is distributed as a Dirichlet random variable, $\bm{\pi} \sim \text{Dir}(\alpha_0)$, with a concentration parameter $\alpha_0$.

The data likelihood for each segment is:
\begin{equation}
\begin{split}
     p(\bm{Y^d_s} &|\bm{Z^d_s},\bm{X^d_s}, \bm{\theta}, \beta) \\ 
     &= \prod_{m=1}^M\mathcal{N}(\bm{Y^d_s}|0, K_{\bm{\theta}_m}(\bm{X^d_s}, \bm{X^d_s}) + \beta \mathbf{I})^{z^d_{s,m}}
\end{split}
\end{equation} with $p(\bm{\theta})= \prod_{m=1}^M p(\bm{\theta_m})$, where $\bm{\theta_m} = (a^2_m, l^2_m)$ are the kernel amplitude and length-scale parameters. We place log-normal prior distributions on the amplitude and the length-scale of the kernel as well as the noise variance $\beta$.


\section{Inference}

We propose to perform inference in the model with a hybrid expectation maximization approach, using Gibbs sampling for the segmentation $\bm{C}$ and a variational EM for the inference over $\bm{Z}$ and ${\bm{\pi}}$. We use a MAP estimate for the kernel parameters $\bm{\theta}_m$ and the noise $\beta$.

\textbf{Fixed hyperparameters}. The following hyperparameters are fixed a priori: the prior on the length of each segment, $\lambda$, the concentration parameter of a symmetrical Dirichlet prior, $\alpha_0$, and the parameters of the log-normal priors on the kernel parameters and the noise.

The algorithm then iterates over the following steps:
\begin{enumerate}
    \item Gibbs sampling over splits $\bm{C}$, given MAP estimates of the parameters $\bm{\theta}^*,\beta^*$.
    \item Variational EM given sampled splits. Variational distribution over cluster assignments $\bm{Z}$ and its mixing parameters $\bm{\pi}$.
    \begin{enumerate}
        \item optimal $q^*(\bm{Z})$, given $q(\bm{\pi})$ and $\bm{\theta}^*,\beta^*$,
        \item optimal $q^*(\bm{\pi})$, given $q^*(\bm{Z})$ and $\bm{\theta}^*,\beta^*$,
    \end{enumerate}
    \item update MAP estimates $\bm{\theta}^*,\beta^*$, given $q^*(\bm{C})$, $q^*(\bm{\pi})$, $q^*(\bm{Z})$, by maximizing ELBO.
    
\end{enumerate}
We now look at each of these steps one at a time.
 
\subsection{Gibbs sampling of segmentation}
\label{gibbs}
It is not possible to analytically marginalise out the segmentation from our model, but we can sample from it using Gibbs sampling. At every time step the probability of splitting into 2 segments only depends on the the 2 adjacent splits, the data in the region between those splits, and the latent cluster assignment of all other segments. Let us denote by $a$ and $b$ the indices of the segments to the left and to the right of each split provided that a split is created at $c^d_i$. We use $a \cup b$ to indicate the union of the segments. Then
\begin{equation}
\label{eq:gibbs}
    \begin{split}
   p(&c^d_i |\bm{Y^d}, \bm{c^d_{-i}},\bm{\theta}, \bm{Z}) \propto p(c^d_i, \bm{Y^d}| \bm{c^d_{-i}},\bm{\theta}, \bm{Z}) \\
    & = \begin{cases}
    p(\bm{Y^d_{a\cup b}}|\bm{\theta}, \bm{Z_{-a\cup b}})p(l_a+l_b), c_i^d=0\\
    p(\bm{Y^d_a}|\bm{\theta}, \bm{Z_{-a}})p(\bm{Y^d_b}|\bm{\theta}, \bm{Z_{-b}})p(l_a)p(l_b), c_i^d=1.
   \end{cases}
   \end{split}{}
\end{equation}

Computing the probability of a particular segment of data, $Y_s$, requires marginalizing over the kernel assignments $\bm{z_s}$ and the mixing distribution $\bm{\pi}$. The latter causes a problem by creating dependencies between the cluster assignments of all segments:
\begin{equation}
\begin{split}
&p(\bm{Y}|\bm{C},\bm{\theta}) = \int_{\bm{\pi}}\prod_{s}\sum_{\bm{z_s}}p(\bm{Y_s}|\bm{z_s}, \bm{\theta}) p(\bm{z_s}|\bm{\pi}) p(\bm{\pi})d\bm{\pi}.
\end{split}
\end{equation}
To address this issue, we marginalize over $\bm{\pi}$ approximately with $p(\bm{\pi}|\bm{Z_{-s}}) \approx q^{*}(\bm{\pi})$, where $q^{*}(\bm{\pi})$ is the optimal approximate posterior given previous segmentation samples computed in the vEM step. During the first iteration, $q^{*}(\bm{\pi})$ is equal to the prior, \emph{i.e.} $q^{*}(\bm{\pi})=p(\bm{\pi})$. Then the likelihood for each segment is:
\begin{equation}
\begin{split}
\label{eq:approx_qc}
 p(\bm{Y_s}&|\bm{\theta}, \bm{Z_{-s}}) \\
& = \sum_{\bm{z_s}}p(\bm{Y_s}|\bm{z_s}, \bm{\theta}) \int_{\bm{\pi}}p(\bm{z_s}|\bm{\pi}) p(\bm{\pi}|\bm{Z_{-s}})d\bm{\pi} \\
& \approx \sum_{\bm{z_s}}p(\bm{Y_s}|\bm{z_s}, \bm{\theta}) \int_{\bm{\pi}} p(\bm{z_s}|\bm{\pi}) q^{*}(\bm{\pi})d\bm{\pi} \\
& = \sum_{m=1}^M p(\bm{Y_s}|\bm{\theta_m}) \mathbb{E}_{q^{*}(\bm{\pi})}[ \pi_m ] =: \Tilde{p}(\bm{Y_s}|\bm{\theta}).
\end{split}
\end{equation}

Using this approximation, $\Tilde{p}(\bm{Y_s}|\bm{\theta})$, to the marginal segment likelihood, we can sample from $\Tilde{p}(\bm{C}|\bm{Y_s},\bm{\theta})\approx p(\bm{C}|\bm{Y_s},\bm{\theta})$ according to \eqref{eq:gibbs}.

\subsection{Gibbs EM}
Recall, that the overall goal is to maximize the marginal likelihood:
\begin{equation}
    p(\bm{Y}|\bm{\theta}, \beta) = \sum_{C}{p(\bm{Y},\bm{C},|\bm{\theta}, \beta)}.
\end{equation}



Following the general EM scheme, we first keep $\bm{\theta}^{old}$ and $\bm{\beta}^{old}$ fixed and optimize $q(\bm{C})$.  For that, we choose a variational distribution over splits to be $q(\bm{C})=\Tilde{p}(\bm{C}|\bm{Y},\bm{\theta}^{old}, \beta^{old})$ (see \eqref{eq:approx_qc}), which corresponds to an approximation to the current estimate of the true posterior given the current kernel and noise parameters.
We can not write it down analytically, and use Gibbs sampling to compute the expectations. The details of the sampling are described in Sec.~\ref{gibbs}.

The evidence lower bound (ELBO) is then:





\begin{equation}
\begin{aligned}
    &\ln{p(\bm{Y}|\bm{\theta}, \beta)} \geq \mathcal{L}(\bm{\theta}, \beta) = 
    \mathbb{E}_{q(\bm{C})}\Big[\ln{\frac{p(\bm{Y},\bm{C}|\bm{\theta}, \beta)}{q(\bm{C})}}\Big] \\
    & = \mathbb{E}_{q(\bm{C})}\Big[\ln{p(\bm{Y}|\bm{C}, \bm{\theta}, \beta)}\Big] - KL(q(\bm{C})||p(\bm{C})) \\
    & \approx \frac{1}{L}\sum_{i=1}^L\Big[\ln{p(\bm{Y}|\bm{C}_i, \bm{\theta}, \beta)}\Big] + \text{const},
\end{aligned}
\end{equation}
where $\bm{C}_i \sim \Tilde{p}(\bm{C}|\bm{Y},\bm{\theta}^{old})$ are samples.

\subsection{Variational EM}

We now need to maximize $\ln{p(\bm{Y}|\bm{C}_i,\bm{\theta}, \beta)}$, given a sample of the segmentation:
\begin{equation}
    p(\bm{Y}|\bm{\theta}, \beta) = \int_{\bm{\pi}}\sum_{\bm{Z}} p(\bm{Y},\bm{Z},\bm{\pi}|\bm{\theta}, \beta) d\bm{\pi}.
\end{equation}
In this section we will omit the dependency on sampled segmentation $\bm{C}_i$ for clarity. Here we use variational EM with MAP estimates for parameters $\bm{\theta}$ and $\beta$, similarly to~\cite{Bishop:2006}.

{\bf E step}
Using the classical result for VI,
\begin{equation}
    \ln{q^*(Z_i)} \propto \mathbb{E}_{q^*(Z_{-i})}\Big[\ln{p(Y,Z|\theta)}\Big],
\end{equation} the optimal distributions $q^*(\bm{Z})$ and $q^*(\bm{\pi})$ can be computed analytically.

Given a sample $\bm{C}_i$, $q(\bm{Z})$  can be factorized as follows $q(\bm{Z})=\prod_{s=1}^{S(C_i)}q_s(\bm{Z}_s)$.
Then the optimal $q^*(\bm{Z})$ is:

\begin{equation}
\begin{aligned}
    &\ln{q^*(\bm{Z})} \propto \mathbb{E}_{q(\bm{\pi})}\Big[\ln{p(\bm{Y},\bm{Z},\bm{\pi}|\bm{\theta}, \beta)}\Big] \\
    & = \sum_{d=1}^D\sum_{s=1}^{S_d}\sum_{m=1}^M z_{d,s}^m\Big[\mathbb{E}_{q(\bm{\pi})}\big[\ln{\pi_m}\big] + \ln{p(\bm{Y^d_s}|\bm{\theta}_m, \beta)}\Big] \\ &+ \text{const}
\end{aligned}
\end{equation}
From this we can see that $q^*(\bm{Z})$ is a categorical distribution, factorized over sequences and segments:
\begin{align}
    &q^*(\bm{Z}) = \prod_{d=1}^D\prod_{s=1}^{S_d}\prod_{m=1}^M (r_{d,s}^m)^{z_{d,s}^m},
\end{align}
where $r_{d,s}^m = \frac{\rho_{d,s}^m}{\sum_{m}\rho_{d,s}^m}$ and  $\rho_{d,s}^m = \exp\{\mathbb{E}_{q(\bm{\pi})}\big[\ln{\pi_m}\big] + \ln{p(\bm{Y^d_s}|\bm{\theta}_m, \beta)}\}$.

While the distribution of the optimal cluster assignments factorises over all segments given $q(\bm{\pi})$, and each sample of segmentation has its own set of segments, the optimal $q^*(\bm{\pi})$ will depend on all samples of $\bm{C}$:
\begin{equation}
\begin{aligned}
    \ln{q^*(\bm {\pi})}& \propto \mathbb{E}_{q(\bm{C}) q^*(\bm{Z})}\Big[\ln{p(\bm{Y},\bm{Z},\bm{\pi}|\bm{C},\bm{\theta}, \beta)}\Big] \\
    & \propto (\alpha_0 - 1)\sum_{m=1}^M \ln{\pi_m} \\
    & + \frac{1}{L}\sum_{i=1}^L\sum_{d=1}^D\sum_{s=1}^{S(C^d_i)}\sum_{m=1}^M \mathbb{E}[z_{d,s,i}^m]\ln{\pi_m} \\
    & = \ln{\prod_{m=1}^M \pi_m^{\alpha_0 - 1 + 1/L\sum_i\sum_d\sum_s r_{d,s,i}^m}}
\end{aligned}    
\end{equation}

Taking the exponentiation of both sides, we can recognize a Dirichlet distribution $q^*(\bm{\pi})=Dir(\bm{\alpha})$, where $\alpha_m = \alpha_0 + \frac{1}{L}\sum_i\sum_d\sum_s r_{d,s,i}^m$.

{\bf M step}
Keeping variational distributions fixed, we find the MAP estimates of the kernel parameters $\bm{\theta}$ and noise  $\beta$:
\begin{equation}
\begin{split}
    \mathcal{L}(\bm{\theta}, \beta) &= \frac{1}{L} \sum_{i=1}^L\sum_{\bm{Z}_i} q^*(\bm{Z}_i)\ln{p(\bm{Y}|\bm{Z}_i,\bm{\theta}, \beta)} \\
    & + \ln{p(\bm{\theta})} + \ln{p(\beta)} + \text{const},
\end{split}
\end{equation}{}
where $\bm{Z}_i$ refers to $\bm{Z}$ for the $i$-th sample of splits $\bm{C}$.





\section{Experiments}

The overall model performance and sensitivity to some extreme cases of segment lengths are tested on synthetic data. We also test the model on real data of heart beat sounds and evaluate the quality of the resulting segmentation on the task of classifying heart sound.

\subsection{Synthetic data}
We first test the inference on synthetic data where we set the noise to be $\beta=0.001$ and the number of kernels to be $M=3$. The kernel parameters are shown in Table \ref{table:synth_1} (where we use SE kernels). We sample 3 sequences with lengths 30, 16, and 20.

During inference the number of available kernels is set to $M = 5$. A Dirichlet prior on the probability of cluster assignments is skewed to promote sparsity ($\alpha_0=0.1$).

\begin{figure}[t]
\centering
\includegraphics[width=\linewidth]{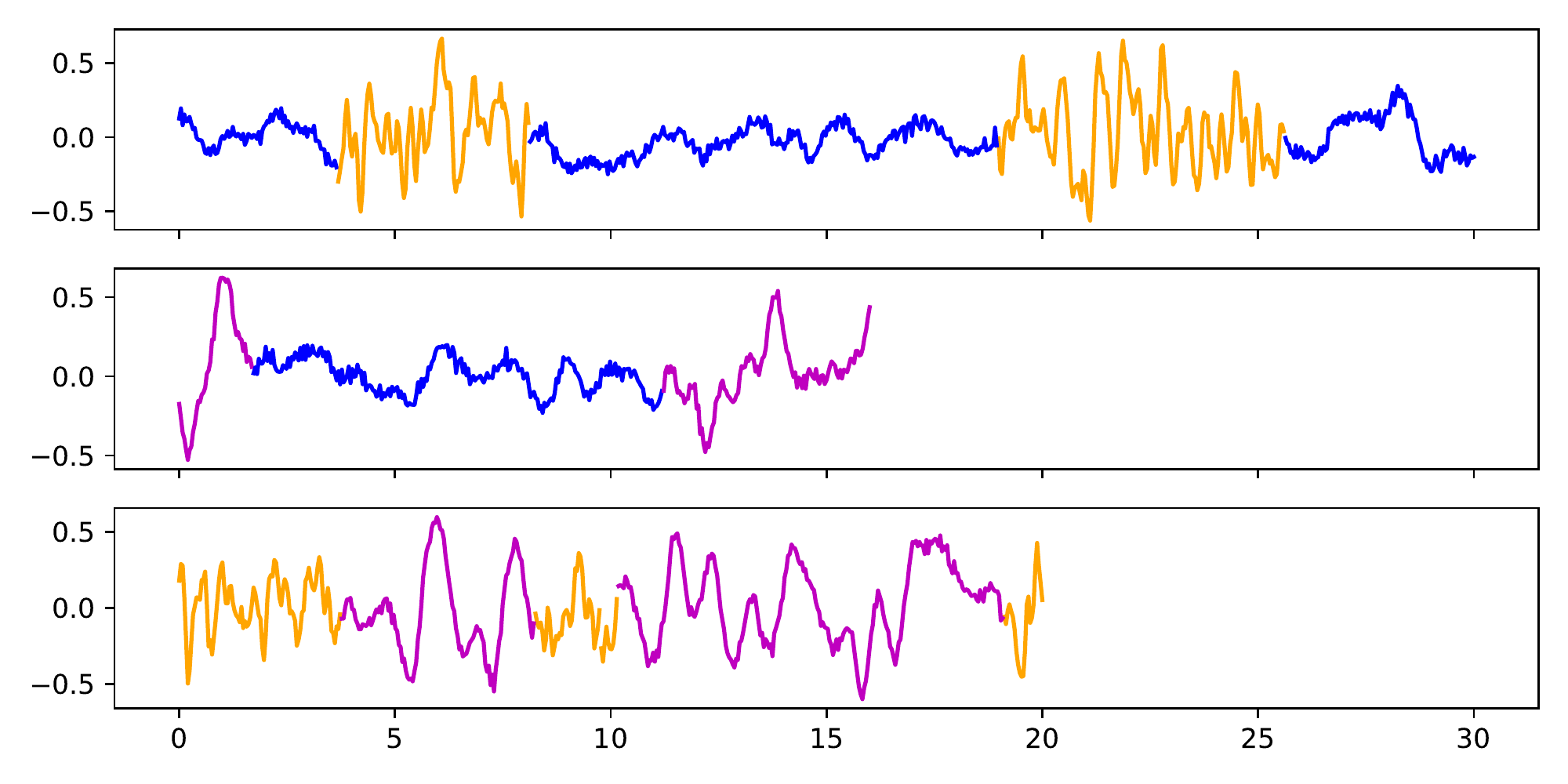}
\vspace{-3mm}
\caption{Segmentation on Synthetic Data. Colors correspond to different kernels}
\label{fig:synthetic}
\end{figure}

\begin{figure}[t]
\centering
\includegraphics[width=\linewidth]{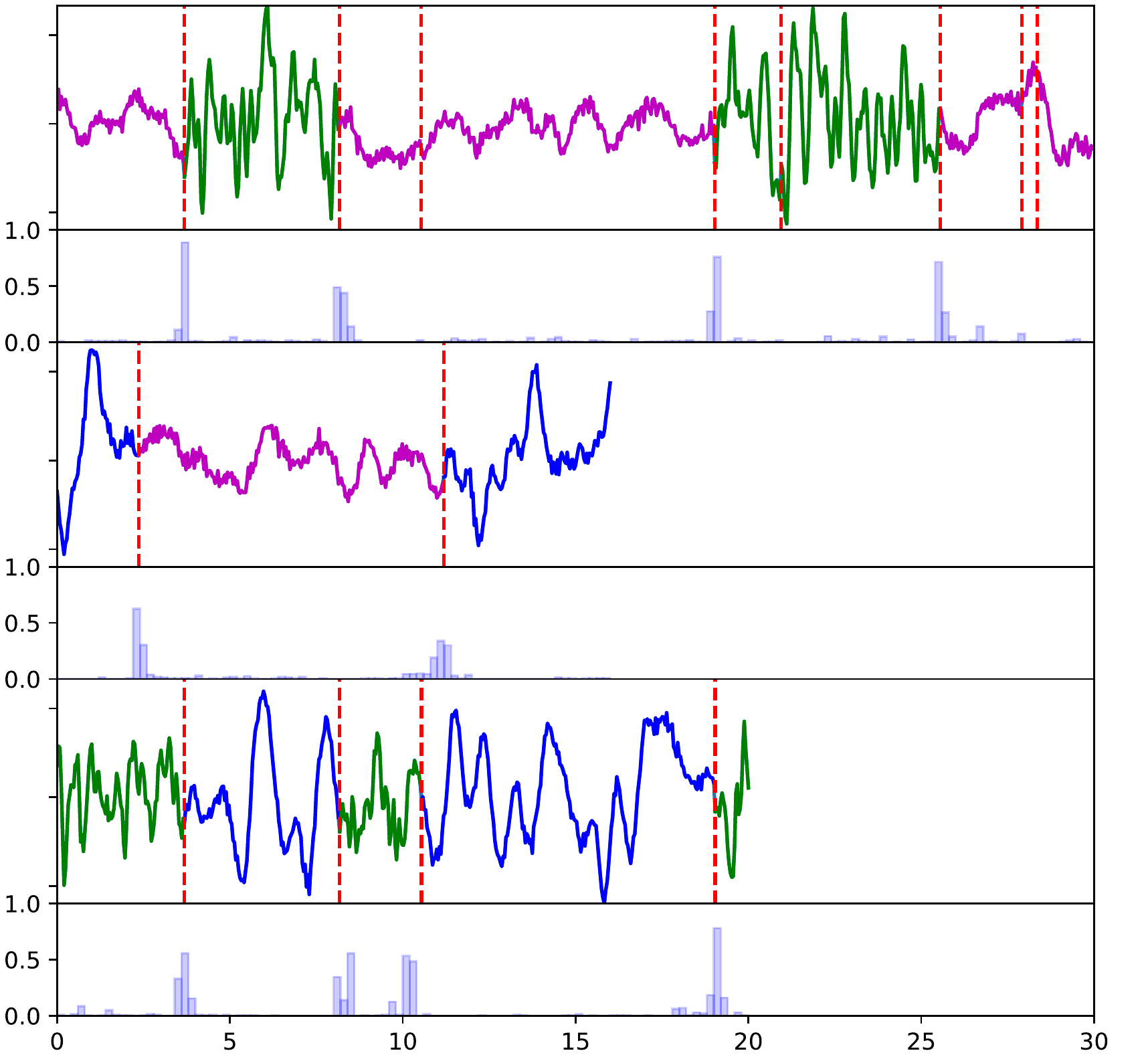}
\vspace{-3mm}
\caption{Ground truth synthetic data and uncertainty over marginal probability of splits. Estimated using 100 samples. Color coding corresponds to the ground truth cluster assignments. Vertical lines are ground truth splits.}
\label{fig:synthetic_uncertainty}
\end{figure}

An example of a segmentation is shown in Figure \ref{fig:synthetic}. 
We illustrate marginal posterior probability of splits in Figure \ref{fig:synthetic_uncertainty}, together with the ground truth segmentation and cluster assignments. The approximate posterior $q(\bm{\pi})$ indicates that only 3 kernels are used in the model and 2 are switched off. Learned MAP estimates for kernel parameters are close to the generating parameters, as we can see from Table \ref{table:synth_1}, where learnt parameters are compared with the most similar generating kernels. So, the model correctly recovers the number of kernel as well as their parameters. The MAP estimate of the noise $\beta^{MAP}=0.0093$ is also close to the generating parameter. 

The 
probability densities over marginal breaks 
correspond to the true segmentation. The only exceptions are splits those where both adjacent segments belong to the same kernel and are by chance seem correlated.


\begin{table}[h]
\centering
\caption{Parameters for Synthetic Data}
\begin{adjustbox}{max width=0.47\textwidth}
\begin{tabular}{|l|r|r|r|r|}
\hline
 & \multicolumn{2}{l|}{\textbf{Ground truth}} & \multicolumn{2}{l|}{\textbf{Learnt}}\\
 \cline{2-5}
 & $a^2$ & $l^2$ & $a^2$ & $l^2$\\
 \hline
1  & 0.01 & 0.1  & 0.0093 & 0.0967 \\
\hline
2  & 0.05 & 0.05  & 0.0513 & 0.0503\\
\hline
3  & 0.05 & 0.005  & 0.0451 & 0.0053 \\
\hline
\end{tabular}
\label{table:synth_1}
\end{adjustbox}
\end{table}

\subsubsection{Edge cases}

We consider 2 edge cases to test model sensitivity to prior. In first case, we have a single segment of the full sequence length. For the second case, the same time period is split into 15 segments of equal lengths. For both cases we only have 1 SE kernel with amplitude $a^2 = 0.1$, and length-scale $l^2=0.4$. Noise is set to $\beta=0.001$.

During the inference maximum number of available kernels is set to $M=5$, Dirichlet prior is $\alpha_0=0.1$. Exponential distribution with $\lambda=0.25$ is used as a prior for the segment lengths, hence implying the mean of 4 (the total sequence length is set to 30).

In both cases inference correctly detects, that only 1 kernel should be used. The comparison of the ground truth and learnt parameters is presented in Table \ref{table:synth_edge}.

\begin{table}[h]
\centering
\caption{Model Parameters for Edge Cases}
\begin{adjustbox}{max width=0.47\textwidth}
\begin{tabular}{|l|r|r|r|}
\hline
 & \textbf{Generating} & \textbf{Edge 1} & \textbf{Edge 2}\\
 \hline
$\beta$  & 0.001 & 0.001 & 0.001 \\
\hline
$a^2$  & 0.1 & 0.061 & 0.081 \\
\hline
$l^2$  & 0.4 & 0.367 & 0.368 \\
\hline
\end{tabular}
\label{table:synth_edge}
\end{adjustbox}
\end{table}

\begin{figure}[h]
\centering
\begin{subfigure}[b]{\linewidth}
\includegraphics[width=\linewidth]{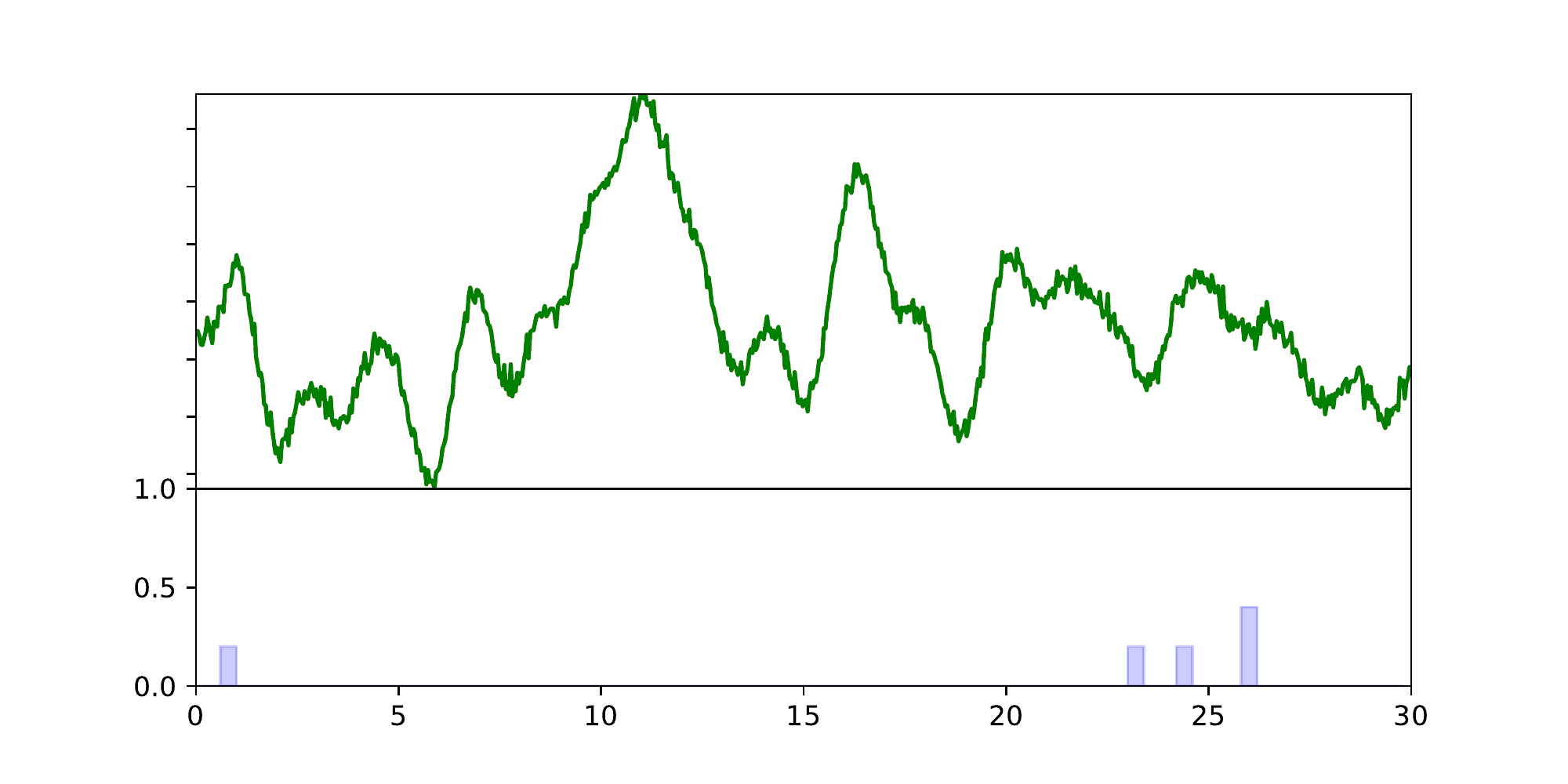}
\vspace{-3mm}
\caption{No breaks (one segment)}
\label{fig:synthetic2}
\end{subfigure}
\begin{subfigure}[b]{\linewidth}
\includegraphics[width=\linewidth]{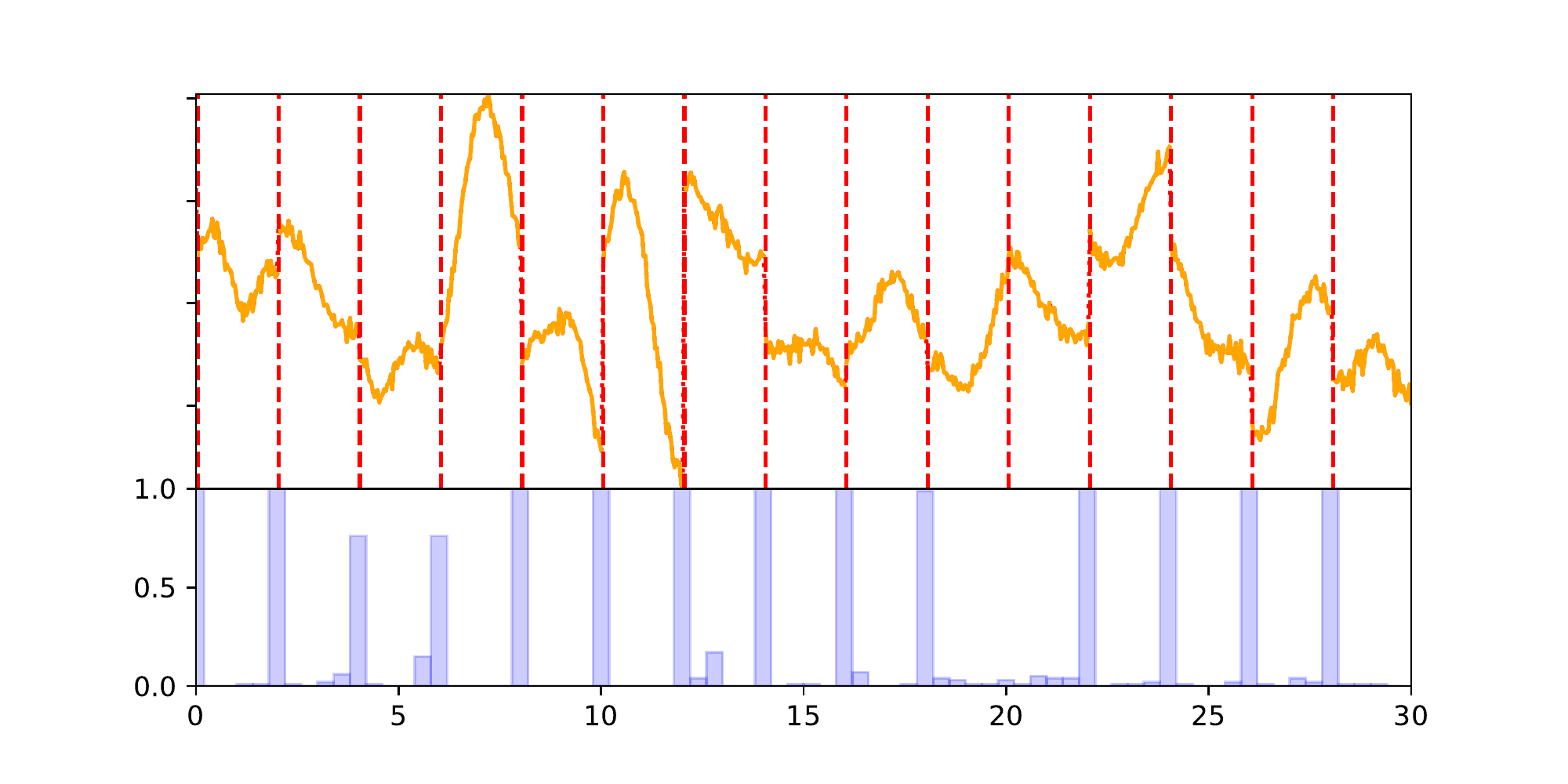}
\vspace{-3mm}
\caption{Identical small length for all segments}
\label{fig:synthetic3}
\end{subfigure}
\caption{Ground truth and split uncertainty for edge-case synthetic data}
\label{fig:synthetic_edge}
\end{figure}

The plots of data, together with ground truth segmentation and marginal uncertainty over splits are shown in Figures \ref{fig:synthetic2} and \ref{fig:synthetic3}.

As we can see from this results, in both extreme cases the model correctly finds the number of kernels and their parameters. Segmentation is also very close to the ground truth. Hence, memoryless Exponential distribution is a reasonable default prior, which leaves most of the segmentation to GP likelihoods.

\subsection{Heart beat sound data}

Segmentation and clustering of sequences gives us a lower dimensional representation of the sequences. We test the quality of the representation on a data set used in "PASCAL Classifying Heart Sounds Challenge"\footnote{http://www.peterjbentley.com/heartchallenge/} (\cite{gomes2013classifying}). The data set is recorded during clinic trial in hospitals using a digital stethoscope. One of the tasks in this challenge, that we choose to tackle, is classification of recorded heart beat sounds (sequences) into one of 3 classes: "normal", "murmur" and "extrasystole". First class contains healthy heart beats with familiar "lub/dub" pattern. The "murmur" class is characterized by additional “whooshing or roaring” noise. Extrasystole sounds may appear occasionally and can be identified because there is a heart sound that is out of rhythm involving extra or skipped heartbeats.

\begin{figure}[t]
\centering
\includegraphics[width=\linewidth]{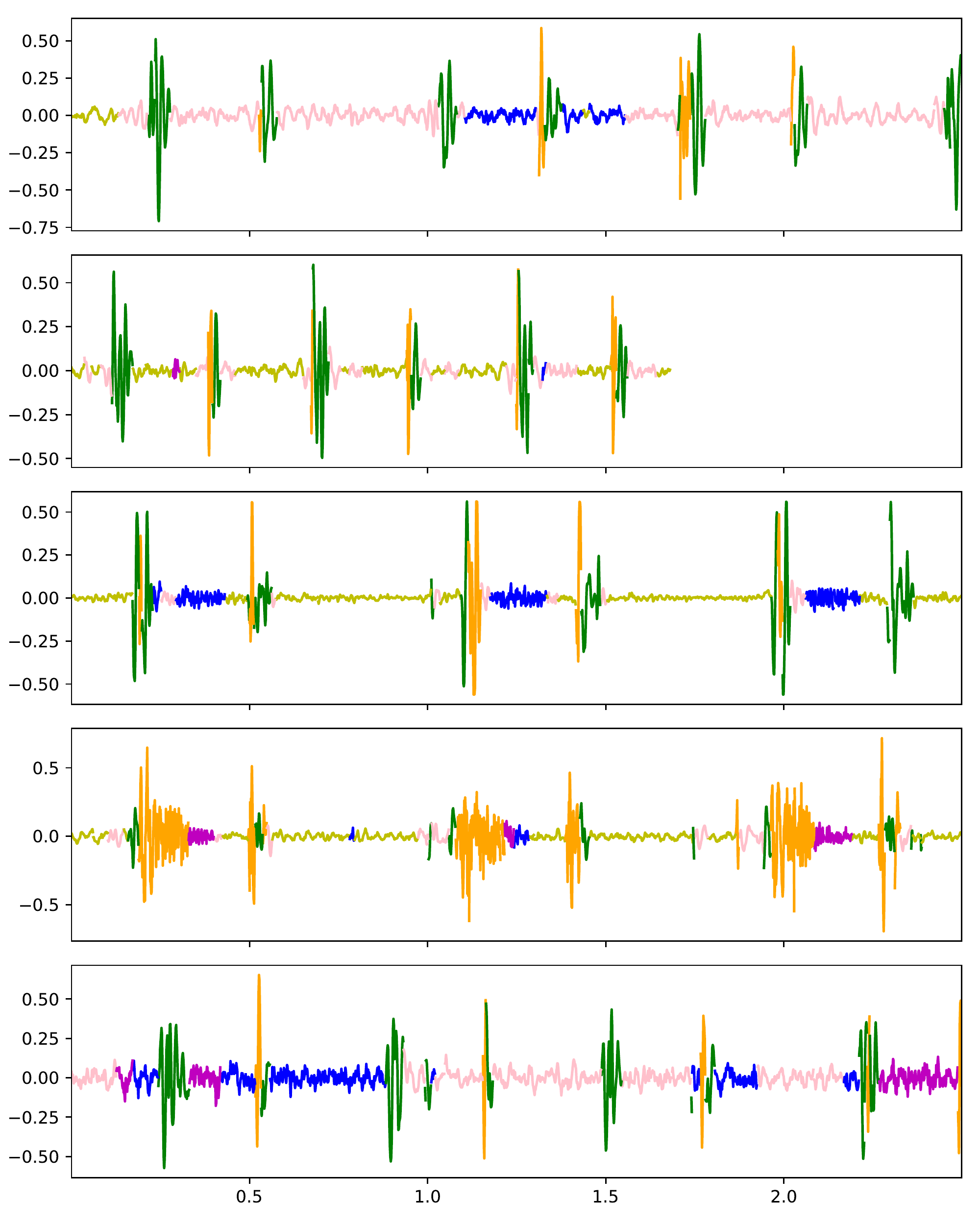}
\caption{Segmentation of part of the heart beats data set, used for training the model. One sample of segmentation is shown. Color coding corresponds to cluster assignment.}
\label{fig:heart_beats_train}
\end{figure}

First, the model is trained on 5 sequences (2 normal, 2 murmur and 1 extrasystole). The number of available kernels is fixed to $M=6$, and put a Dirichlet prior on mixing probability with $\alpha_0=0.1$ to promote switching off unnecessary kernels. The resulting train segmentation is shown in Figure \ref{fig:heart_beats_train}.

We do not include any domain-specific prior knowledge in the model, however in such structured data meaningful prior information would definitely help.

\begin{figure}[t]
\centering
\includegraphics[width=\linewidth]{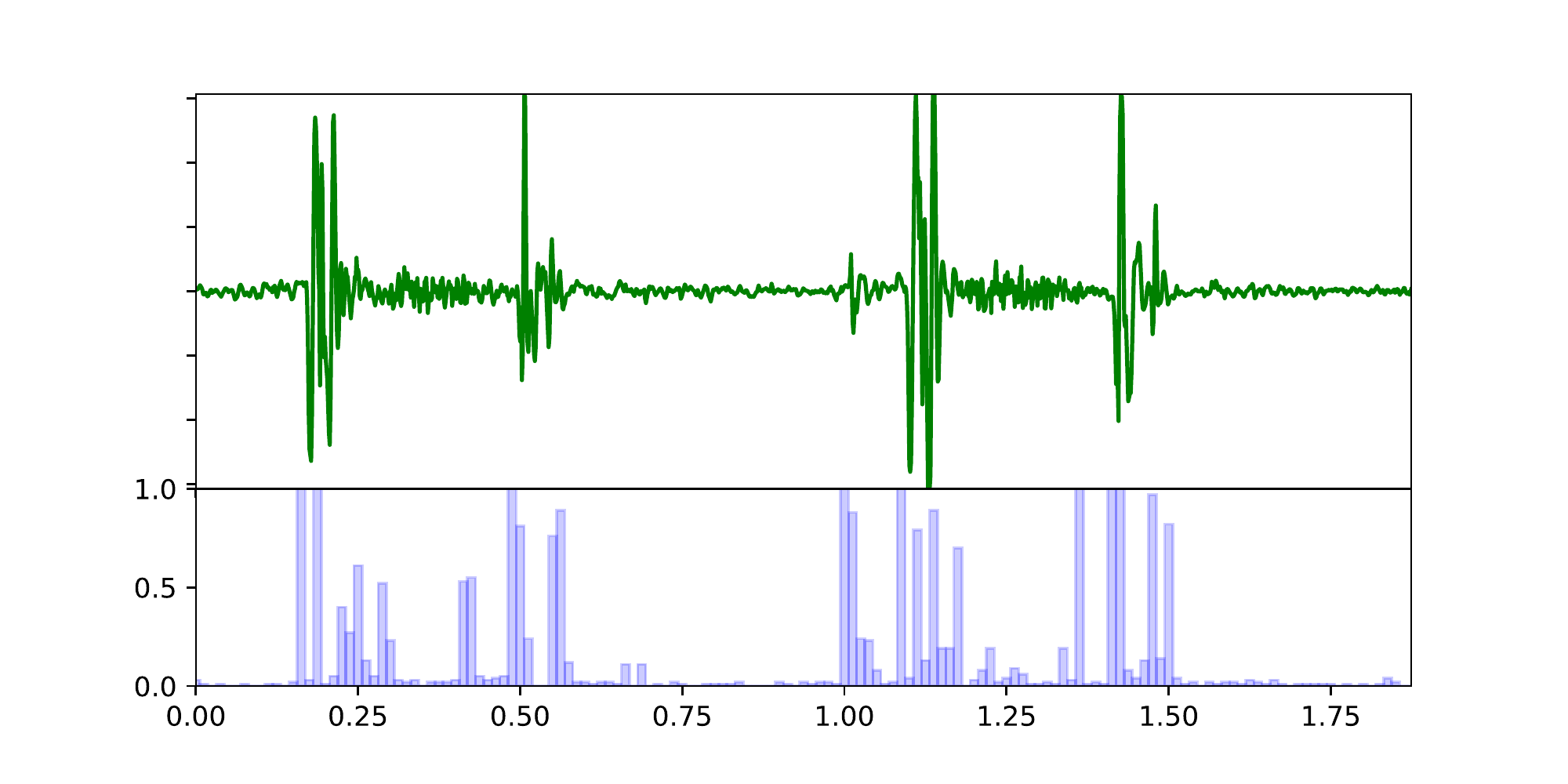}
\vspace{-3mm}
\caption{Uncertainty over marginal probability of splits on an example of heart beat data.}
\label{fig:hartbeats_uncertainty}
\end{figure}

An example of uncertainty over marginal probability of splits is shown in Figure \ref{fig:hartbeats_uncertainty}.

In the next step, we use the pretrained model to segment the whole data set. The size of the data set is shown in Table \ref{table:heat_beat_dataset}.

\begin{table}[h]
\centering
\caption{Heart Sounds Dataset Size}
\begin{adjustbox}{max width=0.47\textwidth}
\begin{tabular}{|r|r|r|r|}
\hline
 \textbf{Normal} & \textbf{Murmur} & \textbf{ExtraS} & \textbf{Test}\\
 \hline
200  & 66 & 46 & 195 \\
\hline
\end{tabular}
\label{table:heat_beat_dataset}
\end{adjustbox}
\end{table}

The resulting segmentation is used as a lower dimentional representation of sequences. More specifically, we represent each sequence as a string, taking most frequent cluster assignment over a window. We use the resulting string as a representation by itself and classify using Support Vector Machines (SVM) with string kernel. We also summarize the string by the frequency of each cluster and train linear SVM on this representation. In our tests the frequency representation with linear SVM classifier gives slightly better results, than string kernel SVM. The results are presented in Table  \ref{table:heart_beats_classification}, compared with the best results of the original challenge (\cite{gomes2013classifying}). Metrics used in comparison are the ones used in the original challenge.

Even thought we did not include any domain-specific prior information, our classification results are comparable to the best results of the challenge, and on 4 out of 8 metrics our results are the best. Our method is good at distinguishing "normal" and "murmur" classes, and bad at predicting "extrasystole". This is expected, as from the description of classes and some examples in Figure \ref{fig:heart_beats_train}, we can conclude that "murmur" can be detected by the presence of special "whooshing" noise. Extra systole, on the other hand, is very similar to the normal heart sounds with occasional skipped or extra beats, which makes it hard to detect using just frequencies of segment types. In this case taking sequential information into account could help, hence our attempt to apply string kernels. Why we did not manage to get good classification results using string kernel SVM on this task, it is an important tool that our representation can leverage.

\begin{table}[h]
\centering
\caption{Heart Sound Classification Results}
\begin{adjustbox}{max width=0.47\textwidth}
\begin{tabular}{|l|r|r|r|r|}
\hline
\multirow{3}{*}{\textbf{}} & \multicolumn{2}{l|}{\textbf{ISEP/IPP}} &
\multirow{2}{*}{\textbf{CS}} &
\multirow{2}{*}{\textbf{Ours}} \\
  & \multicolumn{2}{l|}{\textbf{Portugal}} & & \\ \cline{2-3}
  & \textbf{J48} & \textbf{MLP} &\textbf{UCL} &  \\
 
\hline
Precision of Normal & 0.72 & 0.70 & \textbf{0.77} & 0.74\\
\hline
Precision of Murmur & 0.32 & 0.30 & 0.37 & \textbf{0.65} \\
\hline
Precision of ExtraS & 0.33 & \textbf{0.67} & 0.17 & 0.0\\
\hline
Heart prb Sensitivity & 0.22 & 0.19 & \textbf{0.51} & 0.25\\
\hline
Heart prb Specificity & 0.82 & 0.84 & 0.59 & \textbf{0.94}\\
\hline
Youden Idx Hrt prb & 0.04 & 0.02 & 0.01 & \textbf{0.20}\\
\hline
Discriminant power & 0.05 & 0.04 & 0.09 & \textbf{0.41}\\
\hline
Total precision & 1.37 & \textbf{1.67} & 1.31 & 1.40\\

\hline
\end{tabular}
\label{table:heart_beats_classification}
\end{adjustbox}
\end{table}

\section{Discussion and Future Work}

The proposed model, being probabilistic and generative, naturally allows for tailoring it to a problem at hand by changing the prior distributions. The explicit separation of the segmentation and the cluster assignment steps gives flexible control over the prior probability of the length of the segments, which is hard to do in models where it is merged (for example, Dirichlet process based segmentation~\cite{Dhir:2016}). This construction can be particularly useful, for example, when we know a priori that there is a number of typical lengths. In such case, a mixture of differently located distributions would work well. When using fairly uninformative distributions, like memoryless exponential, we find that the GP priors are more important for the final segmentation.

In contrast to most of the segmentation or change point detection methods, we treat splits as a random variable and do inference over it via Gibbs sampling. This allows for estimation of uncertainty through the marginal probability of the split at each time step. Naturally, uncertainty estimation indicates confidence of the model and can be taken into account in the further decision making process.

While a Dirichlet distribution is used in the model as prior on cluster probability., it can be replaces with a Dirichlet process (DP) to impose a nonparametric prior on clusters. A variational approximation to a DP can be readily incorporated into out inference scheme. However, we find that skewing the Dirichlet distribution towards sparsity, together with appropriate priors on the kernel parameters of the GPs, already gives the desired effect of an automatic determination of the number of clusters.

In the proposed model, the types of the kernels are set a priori but we do not require all kernels have to have the same type. This offers a possible direction for future work, extending the model to include inference over the kernel type.

Although, in this work we consider a fully unsupervised approach, supervision can be easily included in the model. If segmentation if known for some data, it can be used to pretrain the kernels.

One of the benefits of the formulation of our model is that we do not assume temporal alignment of sequences. The dependence of the sequences comes from using the same pool of kernels to generate the data segment-wise. Learning from multiple sequences naturally constrains the problem, and the absence of alignment requirements is beneficial for real-world data collection.

A clear downside of the proposed approach is that the Gibbs sampling component slows the inference in the model (as the sample space grows linearly with the total number of time steps in the data set, and the MC mixing takes more time). This gives a clear direction of future work - improving the inference. Furthermore, while this work focuses on 1-D time-series data, it could be extended to models of multi-dimensional data, another path for future work.

In conclusion, the proposed model allows for interpretability when reasoning about temporal data, by having temporal segmentation as an explicit part of the model. This, for example, allows to map representation features important for classification back to their position in the sequences, while also providing uncertainty. Hence, our approach can be useful in areas, where interpretability is crucial.


\bibliographystyle{abbrvnat}
\bibliography{references}

\end{document}

%% file: graphical_model.tex
\tikzstyle{latent} = [circle,fill=white,draw=black,inner sep=1pt,
minimum size=35pt, font=\fontsize{10}{10}\selectfont, node distance=1]
\tikzstyle{const} = [circle,fill=white,draw=black,inner sep=1pt,
minimum size=15pt, font=\fontsize{10}{10}\selectfont, node distance=1]
\begin{tikzpicture}[scale=1]

\begin{scope}[auto, every node/.style={}]

\node[obs] (y) {$\mathbf{Y^d}$};
\node[obs, left=of y] (x) {$\mathbf{X^d}$};

\node[latent, above=of y] (yl) {$\tilde{\mathbf{Y}}_s^d$};
\node[latent, above=of x] (xl) {$\tilde{\mathbf{X}}_s^d$};

\node[latent, above=of yl, yshift=0.1cm] (z) {$\mathbf{Z}_s^d$};
\node[const, above=of z] (pi) {$\bm{\pi}$};

\node[latent, right=of z] (theta) {$\bm{\theta}_m$};
\node[const, right=of yl] (beta) {$\beta$};
\node[latent, left=of z, xshift=-2cm] (c) {$\bm{C}^d$};

\edge {pi} {z};
\edge {z} {yl};
\edge {yl} {y};
\edge {xl} {yl};
\edge {xl} {x};
\edge {c} {xl};
\edge {c} {yl};
\edge {theta} {yl};
\edge {beta} {yl};

{\tikzset{plate caption/.append style={left=5pt of #1.south west}}
\plate {plate_s} {(z)(xl)(yl)} {$S$} ;}
{\tikzset{plate caption/.append style={below=-8pt of #1.south west}}
\plate {plate_d} {(c)(z)(xl)(yl)(plate_s)(x)(y)} {$D$} ;}

\plate {plate_m} {(theta)} {$M$} ;

\end{scope}
\end{tikzpicture}

%% file: main.bbl
\begin{thebibliography}{14}
\providecommand{\natexlab}[1]{#1}
\providecommand{\url}[1]{\texttt{#1}}
\expandafter\ifx\csname urlstyle\endcsname\relax
  \providecommand{\doi}[1]{doi: #1}\else
  \providecommand{\doi}{doi: \begingroup \urlstyle{rm}\Url}\fi

\bibitem[Aminikhanghahi and Cook(2017)]{Aminikhanghahi:2017}
S.~Aminikhanghahi and D.~J. Cook.
\newblock A survey of methods for time series change point detection.
\newblock \emph{Knowl. Inf. Syst.}, 51\penalty0 (2):\penalty0 339--367, 2017.

\bibitem[Balili et~al.(2015)Balili, Sobrepena, and
  Naval]{balili2015classification}
C.~C. Balili, M.~C.~C. Sobrepena, and P.~C. Naval.
\newblock Classification of heart sounds using discrete and continuous wavelet
  transform and random forests.
\newblock \emph{2015 3rd IAPR Asian Conference on Pattern Recognition (ACPR)},
  pages 655--659, 2015.

\bibitem[Bishop(2006)]{Bishop:2006}
C.~M. Bishop.
\newblock \emph{Pattern Recognition and Machine Learning}.
\newblock Springer, 2006.

\bibitem[Campbell et~al.(2013)Campbell, Liu, Kulis, How, and
  Carin]{Campbell:2013}
T.~Campbell, M.~Liu, B.~Kulis, J.~P. How, and L.~Carin.
\newblock Dynamic clustering via asymptotics of the dependent dirichlet process
  mixture.
\newblock In C.~J.~C. Burges, L.~Bottou, M.~Welling, Z.~Ghahramani, and K.~Q.
  Weinberger, editors, \emph{Advances in Neural Information Processing Systems
  26}, pages 449--457. 2013.

\bibitem[Deng and Bentley(2012)]{deng2012robust}
Y.~Deng and P.~J. Bentley.
\newblock A robust heart sound segmentation and classification algorithm using
  wavelet decomposition and spectrogram.
\newblock In \emph{First PASCAL Heart Challenge Workshop, held after AISTATS},
  2012.

\bibitem[Dhir(2017)]{Dhir:2017:2}
N.~Dhir.
\newblock Bayesian nonparametric methods for dynamics identification and
  segmentation for powered prosthesis control.
\newblock \emph{PhD thesis, University of Oxford}, 2017.

\bibitem[Dhir et~al.(2016)Dhir, Perov, and Wood]{Dhir:2016}
N.~Dhir, Y.~Perov, and F.~Wood.
\newblock Nonparametric bayesian models for unsupervised activity recognition
  and tracking.
\newblock IEEE/RSJ International Conference on Intelligent Robots and Systems
  (IROS), 2016.

\bibitem[Dhir et~al.(2017)Dhir, Wood, Vakar, Markham, Wijers, Trethowan,
  Du~Preez, Loveridge, and Macdonald]{Dhir:2017}
N.~Dhir, F.~Wood, M.~Vakar, A.~Markham, M.~Wijers, P.~Trethowan, B.~Du~Preez,
  A.~Loveridge, and D.~Macdonald.
\newblock Interpreting lion behaviour with nonparametric probabilistic
  programs.
\newblock Association for Uncertainty in Artificial Intelligence, 2017.

\bibitem[Fox et~al.(2008)Fox, Sudderth, Jordan, and Willsky]{Fox:2008}
E.~B. Fox, E.~B. Sudderth, M.~I. Jordan, and A.~S. Willsky.
\newblock An hdp-hmm for systems with state persistence.
\newblock In \emph{Proceedings of the 25th International Conference on Machine
  Learning}, ICML '08, 2008.

\bibitem[Ghosh et~al.(2011)Ghosh, Ungureanu, Sudderth, and
  Blei]{ghosh2011spatial}
S.~Ghosh, A.~B. Ungureanu, E.~B. Sudderth, and D.~M. Blei.
\newblock Spatial distance dependent chinese restaurant processes for image
  segmentation.
\newblock In \emph{Advances in Neural Information Processing Systems}, pages
  1476--1484, 2011.

\bibitem[Gomes et~al.(2013)Gomes, Bentley, Pereira, Coimbra, and
  Deng]{gomes2013classifying}
E.~F. Gomes, P.~J. Bentley, E.~Pereira, M.~T. Coimbra, and Y.~Deng.
\newblock Classifying heart sounds-approaches to the pascal challenge.
\newblock In \emph{HEALTHINF}, pages 337--340, 2013.

\bibitem[Rasmussen and Ghahramani(2002)]{rasmussen2002infinite}
C.~E. Rasmussen and Z.~Ghahramani.
\newblock Infinite mixtures of gaussian process experts.
\newblock In \emph{Advances in neural information processing systems}, pages
  881--888, 2002.

\bibitem[Rasmussen and Williams(2005)]{Rasmussen:2005}
C.~E. Rasmussen and C.~K.~I. Williams.
\newblock \emph{{Gaussian Processes for Machine Learning (Adaptive Computation
  and Machine Learning)}}.
\newblock 2005.

\bibitem[Truong et~al.(2018)Truong, Oudre, and Vayatis]{Truong:2018}
C.~Truong, L.~Oudre, and N.~Vayatis.
\newblock ruptures: change point detection in python, 2018.

\end{thebibliography}
